\documentclass[10pt,journal,compsoc]{IEEEtran}

\ifCLASSOPTIONcompsoc
  \usepackage[nocompress]{cite}
\else
  \usepackage{cite}
\fi

\ifCLASSINFOpdf
\else
\fi

\usepackage{graphicx}
\usepackage{xcolor}
\usepackage{comment}
\usepackage{amsmath,amssymb} %
\usepackage{tabularx}
\newcolumntype{Y}{>{\centering\arraybackslash}X}
\usepackage{booktabs}
\usepackage{multirow}
\newif\showcomments

\newcommand{\etalcite}[1]{ \emph{et al.}~\cite{#1}}
\usepackage{physics}
\usepackage{subcaption}
\usepackage{float}
\usepackage{caption} 
\begin{document}
\title{Multilevel Knowledge Transfer \\for Cross-Domain Object Detection}

\author{Botos~Csaba,
        Xiaojuan~Qi,
        Arslan~Chaudhry,
        Puneet~Dokania,
        Philip~Torr
\IEEEcompsocitemizethanks{
\IEEEcompsocthanksitem Botos Cs. and P. Torr are with the Department of Engineering Science, University of Oxford.
E-mail: \{csbotos, philip.torr\}@robots.ox.ac.uk
\IEEEcompsocthanksitem X. Qi is with the University of Hong Kong.
E-mail: xjqi@eee.hku.hk
\IEEEcompsocthanksitem A. Chaudhry is with DeepMind.
E-mail: arslanch@deepmind.com
\IEEEcompsocthanksitem P. Dokania is with Five AI Ltd. and Department of Engineering Science, University of Oxford.
E-mail: puneet@robots.ox.ac.uk
\IEEEcompsocthanksitem Work was done when P. Dokania and A. Chaudhry were at University of Oxford.
}%
}

\IEEEtitleabstractindextext{%
\begin{abstract}
Domain shift is a well known problem where a model trained on a particular domain (source) does not perform well when exposed to samples from a different domain (target). 
Unsupervised methods that can adapt to domain shift are highly desirable as they allow effective utilization of the source data without requiring additional {\em annotated} training data from the target. 
Practically, obtaining sufficient amount of annotated data from the target domain can be both infeasible and extremely expensive.
In this work, we address the domain shift problem for the object detection task.
Our approach relies on gradually removing the domain shift between the source and the target domains. 
The key ingredients to our approach are -- (a) mapping the source to the target domain on pixel-level; (b) training a teacher network on the mapped source and the {\em unannotated} target domain using adversarial feature alignment; and (c) finally training a student network using the pseudo-labels obtained from the teacher.
Experimentally, when tested on challenging scenarios involving domain shift, we consistently obtain significantly large performance gains over various recent state of the art approaches.

\end{abstract}

}

\maketitle

\IEEEdisplaynontitleabstractindextext

\IEEEpeerreviewmaketitle

\IEEEraisesectionheading{\section{Introduction}\label{sec:intro}}

In the last decade, deep learning has transformed the field of computer vision, becoming the standard approach in the majority of tasks such as classification~\cite{xie2019self}, segmentation~\cite{yuan2019object} and detection~\cite{tan2019efficientdet}. 
In particular, the features learned by deep networks allow for accurate prediction on object detection tasks in a variety of fields such as medical imaging, astronomy, autonomous driving, \emph{etc}.
However, this success comes at a cost of relying on substantial amount of labeled data which may not be available for many practical problems.

Another issue inhibiting the real-world applicability of deep learning based detection algorithms, is their sensitivity to domain shift: when the \textit{target} distribution, from which the test images are sampled, is different than the training \textit{source} distribution. Even the most efficient models suffer severely from this problem~\cite{chen2018domain,saito2019strong,cai2019exploring,khodabandeh2019robust}.
For example, when the weather conditions during test time are different than training time, modern object detectors become unreliable~\cite{sakaridis2018semantic,chen2018domain}.
The naive solution is to collect new training data from the target distribution,
and train a new model from scratch. 
In practice, the high cost of producing object detection labels in particular~\cite{papadopoulos2017extreme,su2012crowdsourcing} renders standard supervised detectors infeasible for general use.

To circumvent the limitations posed by insufficient annotation, Unsupervised Domain Adaptation (UDA) techniques combine the labeled source data with unlabeled samples from the target data distribution.
Following Chen\etalcite{chen2018domain}, growing number of works~\cite{hsu2019progressive,zhu2019adapting,cai2019exploring, saito2019strong,khodabandeh2019robust} utilize UDA and show encouraging results, some even reaching the accuracy of \textit{oracle} models fully trained with labeled target data (see Table~\ref{tab:extensive-comparison}).

\begin{figure*}[t!]
    \centering
    \includegraphics[width=\textwidth]{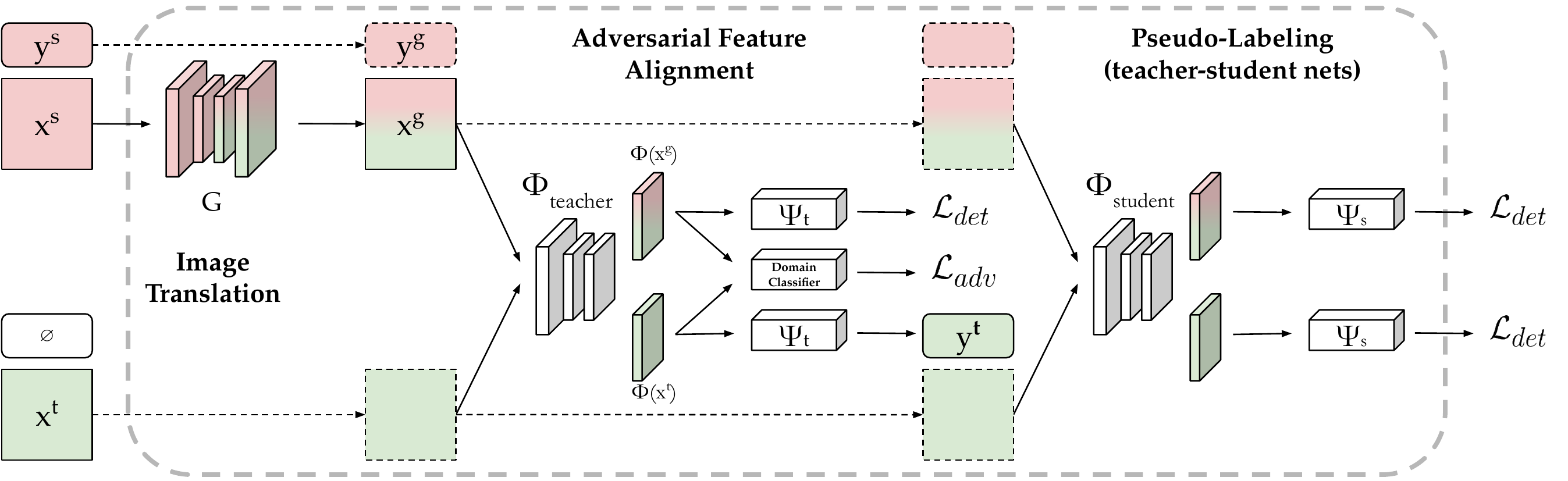}
    \caption{
Multilevel Knowledge Transfer utilizing all three levels of domain adaptation: 1) pixel-level adaptation 2) adversarial feature alignment and 3) pseudo-labeling.
First, the pre-trained image-to-image translation model $G$ converts the labeled source image $x^s$ into a synthetic image $x^g$ that imitates the visual appearance of the images from the target distribution.
Under the assumption that the size and location of the objects are left intact by the translation, we reuse (dashed line) the source labels $y^g=y^s$ for the translated image (see Figure~\ref{fig:visual}).
Second, a teacher network is trained with adversarial feature alignment where the feature extractor $\Phi_{\text{teacher}}$ learns a canonical representation for the synthetic and the target image.
This is achieved by jointly optimizing the original detection loss $\mathcal{L}_{det}$ and adversarially maximizing the cost function of a shallow binary classifier that discriminates features w.r.t their domain.
The latter objective ensures that information about the corresponding domain of the input is eliminated from the learned representation space, while the former is necessary to retain useful features required to carry out the detection task by the predictor head $\Psi$.
Lastly, we use the pre-trained teacher network to generate pseudo-labels $y^t$ to the corresponding target image $x^t$.
The student model is trained without feature level alignment, on the mixture of the labeled synthetic images ($\mathcal{X}^g, \mathcal{Y}^g$) and the pseudo-labeled target images ($\mathcal{X}^t, \mathcal{Y}^t$).
    }
    \label{fig:method}
\end{figure*}

In UDA the domain-specific information can be removed at three levels.
1) \emph{Pixel-level alignment}~\cite{zhu2017unpaired,huang2018auggan}, where the low-level variations such as, color shifts, textures, light conditions etc. are removed between the source and target domains.
2) \emph{Domain-invariant representation learning}~\cite{chen2018domain,he2019multi,zhu2019adapting}, where the model is trained such that it remains invariant to domain-specific information.
3) \emph{Pseudo-labeling}~\cite{cai2019exploring,khodabandeh2019robust}, where a teacher-student network is used to generate additional training data.
Existing works~\cite{hsu2019progressive,kim2019diversify,saito2019strong,inoue2018cross} either improve one of these levels or combine two of them into a joint model.
To the best of our knowledge, none of the existing works jointly study all the three levels for object detection. In this work, we study how each of these three levels progressively remove the domain shift between the source and target domains, and propose a joint model utilizing all three.
Our method (see Figure~\ref{fig:method}) achieves high accuracy, reaching up to 98\% accuracy of the fully-supervised target model, on benchmarks for cross-domain object detection.

Our contributions are as follows:
i) We propose a systematic unification of image-to-image translation, feature alignment and pseudo-labeling into a single training procedure.
ii) Perform ablation study to better understand how various adaptation techniques complement each other to improve the accuracy under various domain shift scenarios.
iii) We evaluate our method on classic unsupervised cross-domain detection settings~\cite{chen2018domain} using the official Faster R-CNN~\cite{ren2015faster} implementation, Detectron2~\cite{wu2019detectron2} and achieve an absolute gain of up to +8.71\% in accuracy over the current state of the art techniques (see Table~\ref{tab:extensive-comparison}).

\section{Related work}

\subsection{Object detection}
Recent state-of-the-art object detection methods can be categorized into two lines of work.
Two-stage detectors that classify and refine region proposals~\cite{girshick2015region,girshick2015fast,ren2015faster,he2017mask} and single-stage detectors~\cite{liu2016ssd,redmon2016you,redmon2017yolo9000,dai2016r} that skip the region proposal stage and directly predict bounding boxes and category scores.
While the latter focuses on decreasing inference time while maintaining accuracy, the former aims to achieve higher accuracy regardless of the speed. An extensive comparison of the two categories can be found in~\cite{huang2017speed}.
For fair comparison with recent methods our work considers Faster-RCNN~\cite{ren2015faster} as baseline, following~\cite{chen2018domain,inoue2018cross,cai2019exploring,hsu2019progressive,zhu2019adapting,saito2019strong,kim2019diversify,khodabandeh2019robust,he2019multi}. \cite{ren2015faster} employs a classic ImageNet~\cite{deng2009imagenet} classifier architecture (e.g. VGG~\cite{simonyan2014very}, ResNet~\cite{he2016deep}, Inception~\cite{szegedy2016rethinking}) as a backbone feature extractor. 
Using the backbone features a Region Proposal Network generates coarse instance predictions that are refined and categorized in the second stage by Region of Interest (ROI) heads. 
Furthermore, Faster-RCNN~\cite{ren2015faster} can be easily extended for other tasks as well, such as keypoint detection, instance segmentation, 3D pose estimation, \textit{etc}.
Our choice of the baseline is also reinforced by the popularity of the reference implementation, Detectron2 by Wu~\textit{et al.}~\cite{wu2019detectron2} which provides a flexible framework and validated evaluation tools, thus facilitates the integration of our method into future works on cross-domain detection.
Although our proposal is evaluated using two-stage detectors, the approach can be easily implemented for single-shot detectors as well.

As pointed out by subsequent works~\cite{chen2018domain,inoue2018cross,hsu2019progressive,zhu2019adapting},
classic detection benchmarks~\cite{pascal-voc-2012,lin2014microsoft,geiger2013vision} only provide limited coverage of real-world challenges,
yet top-notch detectors face severe difficulties when the train and test data distribution differs.
The field that studies such domain-shift scenarios is often referred to as Domain Adaptation.

\subsection{Domain Adaptation}
In the family of Transfer Learning methods, Domain Adaptation (also known as Transductive Transfer Learning) addresses the different-data, same-task problems, while \textit{e.g.} Inductive Transfer Learning focuses on same-data, different-task problems.
A comprehensive overview of DA techniques for visual applications can be found in~\cite{csurka2017comprehensive}, while~\cite{wang2018deep} lists deep neural net adaptation methods.
Our study focuses on Unsupervised Domain Adaptation (UDA) that considers three models:
1) a baseline model trained on supervised source data without adaptation, 
2) a model trained on supervised source data and adapted to unsupervised target data,
3) an oracle trained on supervised target data.
In each case, the accuracy is measured on the target.
The goal is to improve on the baseline accuracy by finding a way to utilize unlabeled target input samples.

While the vast majority of adaptation techniques is mainly focusing on image classification
\cite{panareda2017open,duan2012domain,duan2011visual,fernando2013unsupervised,ganin2014unsupervised,ganin2016domain,ghifary2016deep,gong2012geodesic,gopalan2011domain,kulis2011you,li2017domain,long2015learning,long2016unsupervised,motiian2017unified,sun2016return}, 
recently semantic segmentation
\cite{hoffman2017cycada,zhang2017curriculum,chen2017no,chen2018road,tsai2018learning,zhang2018fully,hong2018conditional} 
and object detection
\cite{huang2018auggan,chen2018domain,inoue2018cross,cai2019exploring,hsu2019progressive,zhu2019adapting,saito2019strong,kim2019diversify,khodabandeh2019robust,he2019multi}
has gained more traction.
Conventional adaptation methods include
Multiple Kernel Learning (MKL)~\cite{gonen2011multiple},
adaptive-MKL~\cite{duan2011visual},
domain transfer-MKL~\cite{duan2012domain},
geodesic flow kernel~\cite{gong2012geodesic},
deformable part based models~\cite{xu2014domain,xu2014incremental},
asymmetric metric learning~\cite{kulis2011you,xu2019deep},
co-variance matrix lignment~\cite{wang2018deep},
alignment of second order statistics~\cite{sun2016return} and
subspace alignment~\cite{raj2015subspace,fernando2013unsupervised}.
In the advent of deep learning, adaptation techniques were refocused on achieving domain-invariant representation during training.
In our work, we categorize these techniques into three classes based on which point of the data-flow do they take effect on, namely: image, feature and output level adaptation methods.

\subsubsection{Image level.}
Formerly known as reweighting algorithms~\cite{shimodaira2000improving,huang2007correcting}, image-level adaptation refers to techniques that transform the labeled source samples to appear as target samples while retaining their semantic content such that the source labels will be still meaningful after the transformation.
Such transformation can be learned in an unsupervised fashion using models inspired by pix2pix~\cite{isola2017image} and Coupled-GAN~\cite{liu2016coupled}, such as CycleGAN~\cite{zhu2017unpaired}, UNIT~\cite{liu2017unsupervised} and MUNIT~\cite{huang2018multimodal}.
Furthermore, there are image translation algorithms specifically tailored for cross-domain adaptation, AugGAN~\cite{huang2018auggan} and SPLAT~\cite{tzeng2018splat}.

\subsubsection{Feature level.}
The goal of feature-level adaptation is to encourage the feature extractor to preserve discriminative features and to discourage learning domain-specific representations.
In~\cite{long2015learning,long2016unsupervised,sun2016deep,cariucci2017autodial} the mean embedding of the source and target distributions are matched using the Maximum Mean Discrepancy (MMD) metric.
In another line of work~\cite{ganin2014unsupervised,ganin2016domain,tzeng2017adversarial} employs adversarial training from~\cite{goodfellow2014generative} to approximate the $\mathcal{H}$-divergence~\cite{chen2018domain} by training a domain classifier network on the learned representations and trains the feature extractor to maximize the error of the classifier.

\subsubsection{Output level.}
Compared to the image level adaptation where one approximates the target \textit{image} distribution, methods that approximate the target \textit{label} distribution are considered output level adaptation methods.
In our work we consider a trivial solution for doing so, which is the standard teacher-student setup~\cite{scudder1965probability} consisting of two stages:
training a teacher network on the source dataset and reuse it to generate pseudo-labels on the target images, then training a student model on the union of labeled and pseudo-labeled data.
Subsequent works following this setup are more commonly studied in the field of semi-supervised learning~\cite{yarowsky1995unsupervised,Laine2016TemporalEF,tarvainen2017mean,inoue2018cross,xie2019self} as well as in Knowledge Distillation~\cite{hinton2015distilling} and Pseudo-Labeling~\cite{lee2013pseudo,iscen2019label,shi2018transductive}.
Sun\etalcite{sun2019unsupervised} suggests a major overlap between the field of domain adaptation and self-supervised learning.
Similarly to image and feature level adaptation, ~\cite{tsai2018learning,tsai2019domain} approaches output level adaptation with adversarial training.
Most of these studies restrict their experiments to image classification or semantic segmentation, however, the findings of~\cite{khodabandeh2019robust,inoue2018cross} show encouraging results in cross-domain object detection.

\subsection{Cross-Domain Object Detection}
\label{subsec:cdod}
Approaches prior to~\cite{chen2018domain} include deformable part-based models~\cite{xu2014domain,xu2014incremental} and subspace alignment methods~\cite{fernando2013unsupervised,raj2015subspace}.
These solutions were limited to specific cases while more general real-world challenges remained to be unsolved.

Chen\etalcite{chen2018domain} suggested three realistic experimental settings in the context of autonomous driving: synthetic-to-real transfer, changing weather conditions and different recording devices.
The source and target datasets are respectively~\cite{johnson2016driving,cordts2016cityscapes},~\cite{cordts2016cityscapes,sakaridis2018semantic} and~\cite{geiger2013vision,cordts2016cityscapes}.
Works after~\cite{chen2018domain} follow these settings to measure the efficiency of their proposed adaptation procedure.

DA-Faster-RCNN~\cite{chen2018domain} adopts adversarial feature alignment~\cite{ganin2014unsupervised} for two-stage object detectors: a domain discriminator is trained on the final stage of the backbone and another discriminator on the box predictor head while the feature extractors are trained to confuse the domain classifiers.

Multi-Adversarial Faster-RCNN (MAF)~\cite{he2019multi} extends~\cite{chen2018domain} to align multiple stages of the backbone feature extractor.

Pseudo-Labeling for object detection is studied in~\cite{inoue2018cross}, which also employs CycleGAN~\cite{zhu2017unpaired} to perform image-level adaptation as a preprocessing step.

MTOR~\cite{cai2019exploring} adopts the Mean Teacher paradigm~\cite{tarvainen2017mean} for object detection by integrating Object Relations into the consistency cost between student and teacher models.

Selective Cross-Domain Alignment (SCDA)~\cite{zhu2019adapting} and Strong-Weak Distribution alignment (SWDA)~\cite{saito2019strong} argues that previous approaches were focusing on ensuring strict alignment on the global representation, despite it can be harmful to tasks heavily relying on spatial information.
SCDA~\cite{zhu2019adapting} breaks down the problem of cross-domain object detection to two sub-problems: \textit{"where to look"} and \textit{"how to align"}, while SWDA~\cite{saito2019strong} implements a different strategy for soft alignment of low-level local features and strict alignment of high-level global features.
To further improve results,~\cite{saito2019strong} uses image-level adaptation in addition to feature alignment in some of their experiments.

Inspired by~\cite{gopalan2011domain}, Progressive Domain Adaptation~\cite{hsu2019progressive} suggests combining image and feature level adaptation in two stages: first adapting a model between source to the intermediate domain constructed by image translation, followed by adaptation between intermediate and target domain.

Diversify and Match (DM)~\cite{kim2019diversify} proposes an alternative approach to integrate image and feature level adaptation by training multiple translation models with different objective functions to map the source domain into various distinctive intermediate domains and trains the detector using Multi-domain-invariant Representation Learning (MRL), a generalization of~\cite{ganin2016domain}.

Robust R-CNN~\cite{khodabandeh2019robust} studies how can one deal effectively with noisy labels to better utilize pseudo-labels generated by the teacher network. 
Their approach augments the teacher-student setup with an external image classifier trained on the source data using hard labels and on the pseudo-labeled target using soft labels. 
The classifier is then used to filter the teacher's predictions and provide soft-labels for training the student model on the target data with the refined pseudo-labels.

The core limitation of approaches that perform only image~\cite{huang2018auggan}, feature~\cite{chen2018domain,zhu2019adapting,he2019multi} or output~\cite{cai2019exploring,khodabandeh2019robust} level adaptation is their narrow focus on improving existing solutions in one particular direction, not considering the merits of other paradigms.
On the other hand,~\cite{hsu2019progressive,saito2019strong,kim2019diversify} successfully combine image-and-feature level or image-and-output level adaptation techniques~\cite{inoue2018cross}, indicating that adaptation methods of different levels complement each other.
From this point of view, one question arises naturally: why not utilize all three levels of adaptation?
In this paper, we propose a method to integrate a simple technique from each level and demonstrate its efficiency, superior to more sophisticated methods with a narrower focus.
\section{Multilevel Knowledge Transfer}
\begin{figure*}[t!]
    \centering
    \includegraphics[width=.85\textwidth]{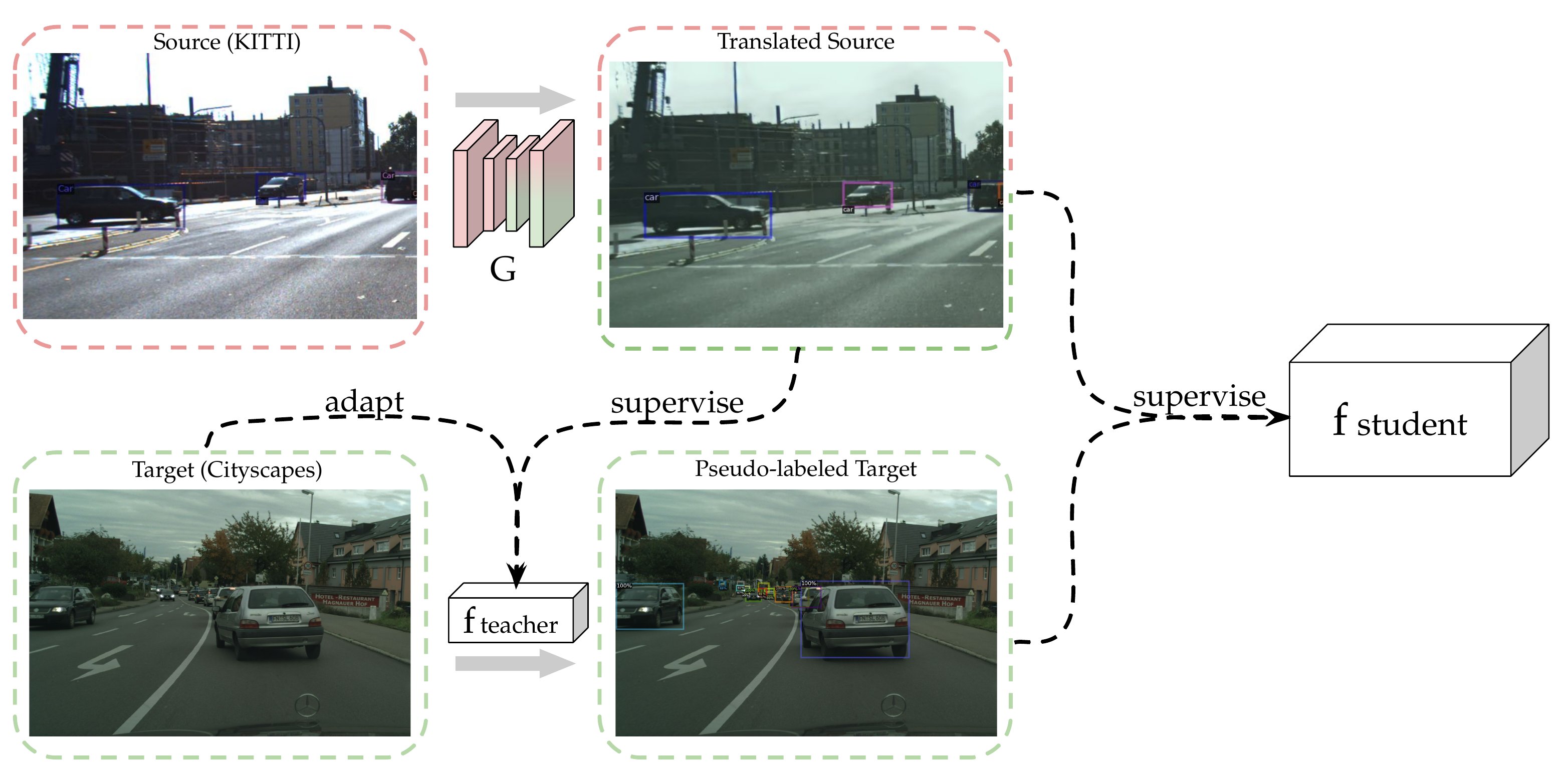}
    \caption{
    Illustration of our method on the \textit{cross-camera} domain adaptation scenario.
    The labeled source data (Cityscpaes~\cite{cordts2016cityscapes}) is transformed by a pre-trained $G$ to make it appear as if it was a sample from the target domain (KITTI~\cite{geiger2013vision}). 
    We enforce $G$ to learn low level visual transformations, thus only textural differences are adopted, whereas the geometric properties of the underlying objects remain intact, therefore the original source labels can be reused.
    The translated source data is then used to provide supervised training signal to the teacher model $f_{teacher}$, meanwhile the unlabeled target data is used in the domain adversarial adaptation of $f_{teacher}$.
    Pseudo-labels are provided for the target data by $f_{teacher}$, which in turn is used in combination with the translated source data to train the final model, $f_{student}$.
    For full details of the process see Figure~\ref{fig:method}.
    }
    \label{fig:visual}
\end{figure*}

\begin{sloppypar}

We now begin the exposition of our method. 
The goal in supervised object detection is to learn a function $f_{\theta}=\Psi \circ \Phi: \mathcal{X}^t \to \mathcal{Y}^t$, parametrized by $\theta$ (a neural network in our case), mapping an input image to the bounding box coordinates and labels of all classes present in the dataset of interest (defined as target data in this work). 
As noted in Sec.~\ref{sec:intro}, in many real-world setups it is difficult to collect a labeled dataset for this task. 
Unsupervised Domain Adaptation (UDA) attempts to use data from another domain, referred to as source domain ($\mathcal{X}^s, \mathcal{Y}^s$), where labeled data is readily available and combines it with unlabeled target data to improve final performance on the target domain test set. 
One way to achieve this in UDA is to gradually remove the domain shift between the source and the target domain, so that both labeled $(\mathcal{X}^s, \mathcal{Y}^s)$ and unlabeled $\mathcal{X}^t$ can be leveraged for the better overall performance on the target domain.

Our proposed model, Multilevel Knowledge Transfer (see Figure~\ref{fig:method}) can be summarized in three steps:
1) {\em Translate} the supervised source images into the target domain.
2) Train a model on the translated images using the source labels with additional feature alignment constraint between the translated source and the target domain.
3) Generate pseudo-labels for the target images from the model obtained in the previous step to train a new model on the combined dataset of translated source dataset and the pseudo-labeled target dataset. Below we describe each step in detail.

\subsection{Image-to-Image translation}
In image level adaptation we remove low-level domain specific variations of the input space, such as: color shifts, textures, light conditions, reflections etc.
For this, we transform the labeled source images $\mathcal{X}^s$ to a synthetic dataset $\mathcal{X}^g$ that approximates the distribution of the target data $\mathcal{X}^t$ by removing the domain shift between the two distributions. We do so by training an image-to-image translation model, Multimodal Unsupervised Image Translation (MUNIT)~\cite{huang2018multimodal}, that requires unlabeled input images from both source and target domains to learn the translation $G:\mathcal{X}^s \to \mathcal{X}^g$.
One notable advantage of MUNIT is its ability to transform a single source image to multiple different synthetic images with identical content but different appearances. Therefore, in addition to reducing the domain shift between the two distributions, MUNIT captures diversity in the target domain. 
This effectively increases the size of the training set by many folds. 
We assume that MUNIT only modifies the appearance, not the geometry, of the original object. The bounding-box class, size and location does not change, therefore, source labels can be reused without modification for the translated images: $\mathcal{Y}^g=\mathcal{Y}^s$.
Prior work~\cite{hsu2019progressive,saito2019strong,kim2019diversify} uses CycleGAN~\cite{zhu2017unpaired} for the same purpose. However, CycleGAN does not guarantee multi-modality resulting in reduced robustness of the model as show in Section~\ref{sup:cyclegan}.

\subsection{Feature level adaptation}
With image-to-image translation covering the dataset aspect of training, we now turn our attention towards the model itself. To increase robustness, we aim to train the model such that learned features ignore domain-specific information and only encode domain-invariant discriminative information. For this, we train a model, referred to as the \emph{teacher}, that \emph{aligns} the features between the target ($\mathcal{X}^t$) and the translated source images ($\mathcal{X}^g$) (obtained in the previous step). We use adversarial training for this alignment. More specifically, the teacher model $f_{t}$ consists of a feature extractor $\Phi_t$ and a predictor $\Psi_t$, making the predictions $f_{t}(x)=\Psi_{t} \circ \Phi_{t} (x)$.
Similarly, we define the student model $f_s(x)=\Psi_s \circ \Phi_s (x)$.
In addition to $\Psi_t$, we train a \textit{Domain Classifier} ($D$) on top of $\Phi_t$ that 
outputs $1$ when the input is in the target domain and $0$ otherwise by minimizing the following loss:
\footnote{Please note, that in our notation \emph{t} as a superscript means \emph{target} (such as $\mathcal{X}^t$), whereas as a subscript it refers to the \emph{teacher} model (e.g. $f_t$)} 

\begin{equation}
    \begin{split}
    \mathcal{L}_{adv}=~
    &\mathbb{E}_{x\sim p(x^{g})}\left[\log D(\Phi_{t}(x))\right] + \\
    &\mathbb{E}_{x\sim p(x^{t})}\left[\log \big( 1-D(\Phi_{t}(x)) \big) \right].    
    \end{split}
\end{equation}
Note that maximizing $\mathcal{L}_{adv}$ over the parameters of $\Phi$ eliminates domain-specific information in the learned feature space, as it maximizes the error of $D$.
Using GRL~\cite{ganin2016domain}, we simultaneously minimize $\mathcal{L}$ over the parameters of $D$ and maximize it over the parameters of $\Phi$.
Finally, the overall training objective $\mathcal{L}_{teacher}$ is given by:
\begin{equation}
    \mathcal{L}_{teacher}=
    \sum_{(\mathcal{X}^g,\mathcal{Y}^g)}\mathcal{L}_{det}(f_{t}(x), y) - 
    \mathcal{L}_{adv}
\end{equation}
where $\mathcal{L}_{det}$ is the standard detection loss~\cite{ren2015faster}.

\subsection{Teacher-student training}
While the domain-invariant model trained in the previous step can be used on the target test set, one can further refine the model by generating \textit{pseudo-labels} $\mathcal{Y}^t$ for the target images and then training a student model on the joint dataset $(\mathcal{X}^g \cup \mathcal{X}^t, \mathcal{Y}^g \cup \mathcal{Y}^t)$. We obtain pseudo-labels by applying the teacher model $f_t$, explained in the previous step, on unlabeled target images, and hard-thresholding the outputs at 0.5 \textit{objectness} score. Experimentally, we found hard-thresholding to be more effective than soft-labeling (see Sec.~\ref{sup:soft-labels} for details). Note, contrary to prior work, such as Noisy Labeling~\cite{khodabandeh2019robust} that uses filtering based on a separate network (43M additional parameters trained for 300k iterations), our method is simple and efficient. 
In addition, in contrast to ~\cite{inoue2018cross}, we do not iterate over multiple teacher-student trainings. 
Once the pseudo-labels from the teacher network are available, the training over the joint dataset $\mathcal{D}=(\{\mathcal{X}^g \cup \mathcal{X}^t\}, \{\mathcal{Y}^g \cup \mathcal{Y}^t\})$ proceed as per the following objective:

\begin{equation}
    \mathcal{L}_{student}=
    \sum_{x,y\in\mathcal{D}}\mathcal{L}_{det}(f_{s}(x), y) 
\end{equation}

An important design choice is that we omit feature level adaptation for the student training.
The reason behind excluding adversarial feature level adaptation is that we have found empirically the supervised training on pseudo-labeled target images sufficient and more stable w.r.t. convergence rate and target accuracy.

\end{sloppypar}
\section{Experiments}

\begin{figure}[ht]
    \centering
    \includegraphics[width=.5\textwidth]{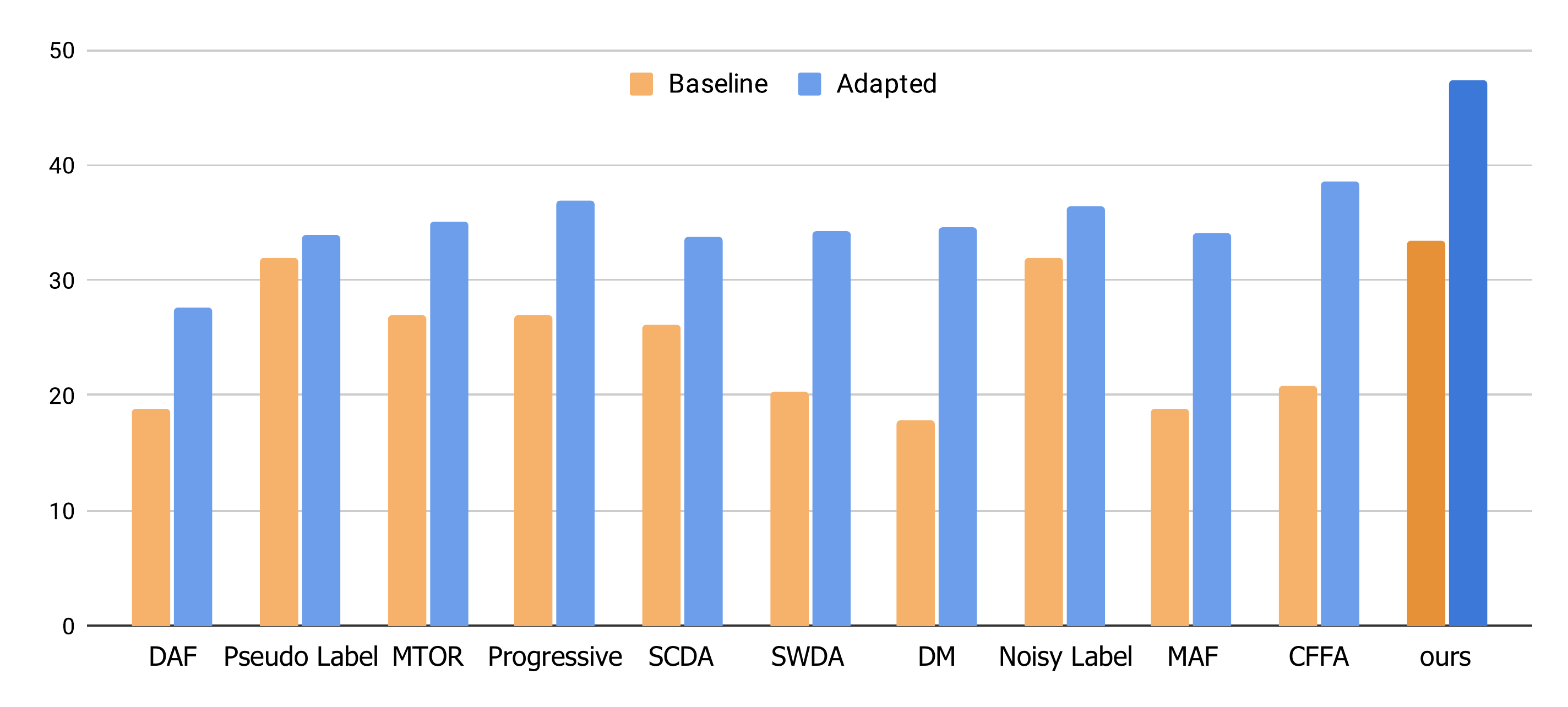}
    \caption{
    Cityscapes$\rightarrow$Foggy Cityscapes mAP\textsubscript{50} scores.
    The bars on the left are the baseline scores reported in prior methods.
    Apart from illustrating the superiority of our approach, we use this figure to illustrate the sensitivity of the mAP\textsubscript{50} metric w.r.t. the underlying implementation.
    }
    \label{fig:cityscapes2foggycityscapes-bars}
\end{figure}

\subsection{Implementation Details}
We adopt ResNet-50~\cite{he2016deep} as our backbone network and follow~\cite{chen2018domain} to set the hyperparameters including training iterations, learning rate policy, batch size and number of proposals in a batch. Our implementation is based on Faster-RCNN~\cite{ren2015faster} in Detectron2~\cite{wu2019detectron2}.
For every experiment we resize the input images such that their shorter side has a length of 600 pixels. Unless stated otherwise, no data augmentation is used. 
Mini-batch of size 2, consisting of one source image with corresponding label and one target image without label, is used for feature level adaptation~\cite{chen2018domain}, and mini-batch of size 1 consisting of a single labeled source image is used when feature level adaptation is not used.
The rest of the hyper-parameters are adapted from \cite{chen2018domain}.
The backbone network is initialized with parameters pre-trained on ImageNet~\cite{deng2009imagenet} and the detection model is trained for 50k iterations using SGD with momentum of 0.9, learning rate of 0.001 and weight decay of 0.0005. The learning rate is then reduced to 0.0001 and the network is further trained for 20k iterations. 

The scores are reported on the validation set of the target domain, after training the model for 70k iterations. 
We report the mean average precision (mAP) with a threshold of 0.5 using the generalized COCO evaluation tool in~\cite{wu2019detectron2}. 
To better evaluate the efficiency of various adaptation techniques, we also propose a new metric that represents to what extent the performance-gap (between source-supervised \textit{baseline} and target-supervised \textit{oracle} models) has been covered by the \textit{adapted} model:
\begin{equation*}
    coverage = \frac{\mathrm{mAP}(adapted) - \mathrm{mAP}(baseline)}{\mathrm{mAP}(oracle) - \mathrm{mAP}(baseline)}
\end{equation*}

\subsection{Datasets}

{Cityscapes}~\cite{cordts2016cityscapes} is a dataset for autonomous driving experiments in urban settings where images were captured with an on-board device.
{SIM 10K}~\cite{johnson2016driving} has 10,000 synthetic images captured from the video game \textit{Grand Theft Auto V}.
{Foggy Cityscapes}~\cite{sakaridis2018semantic} is an artificial dataset  to measure model performance in adverse weather conditions by adding synthetic fog to the real images in Cityscapes~\cite{cordts2016cityscapes}, which contains three different scenarios with decreasing level of objects visibility.
Finally, {KITTI}~\cite{geiger2013vision} provides a dataset for autonomous driving in diverse real-world traffic scenarios ranging from rural areas to inner-city environments.
 Please refer to the Sec.~\ref{sup:datasets} for further details on the datasets.

We follow the cross-domain object detection setups introduced by Chen \textit{et al.}~\cite{chen2018domain}.
Namely, we evaluate our method in \textbf{sim2real} (SIM 10K~\cite{johnson2016driving} $\rightarrow$ Cityscapes~\cite{cordts2016cityscapes}), 
\textbf{changing weather} (Cityscapes~\cite{cordts2016cityscapes} $\rightarrow$ Foggy Cityscapes~\cite{sakaridis2018semantic}) 
and \textbf{cross-camera} adaptation (KITTI~\cite{geiger2013vision} $\rightarrow$ Cityscapes~\cite{cordts2016cityscapes}) settings.

\subsection{Comparison with State-Of-The-Art}
\label{subsec:exp-results}
\begin{figure*}
\begin{subfigure}{.33\textwidth}
  \centering
  \includegraphics[width=\linewidth]{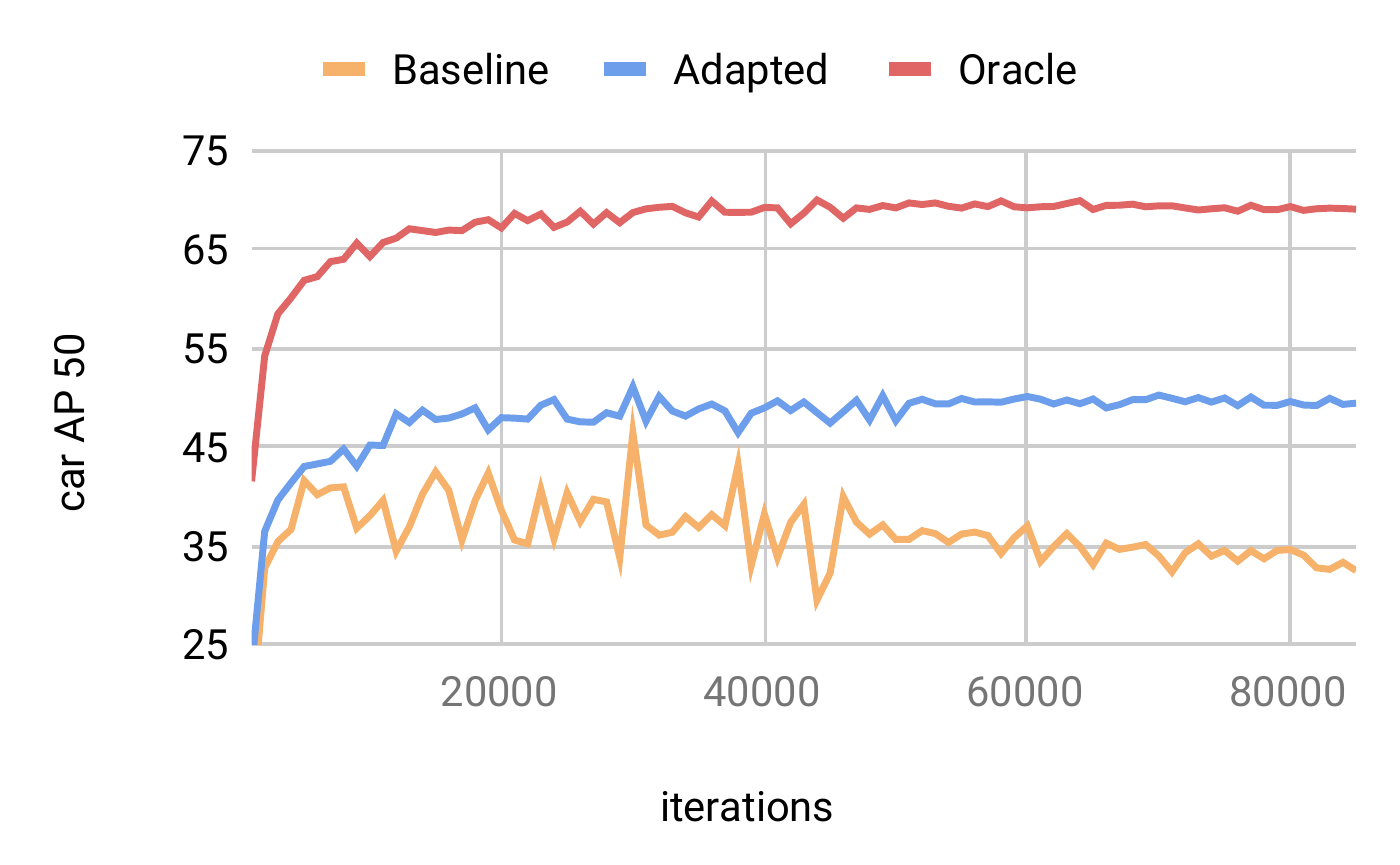}
  \caption{SIM 10K$\rightarrow$Cityscapes}
  \label{fig:sim10k2cityscapes}
\end{subfigure}%
\begin{subfigure}{.33\textwidth}
  \centering
  \includegraphics[width=\linewidth]{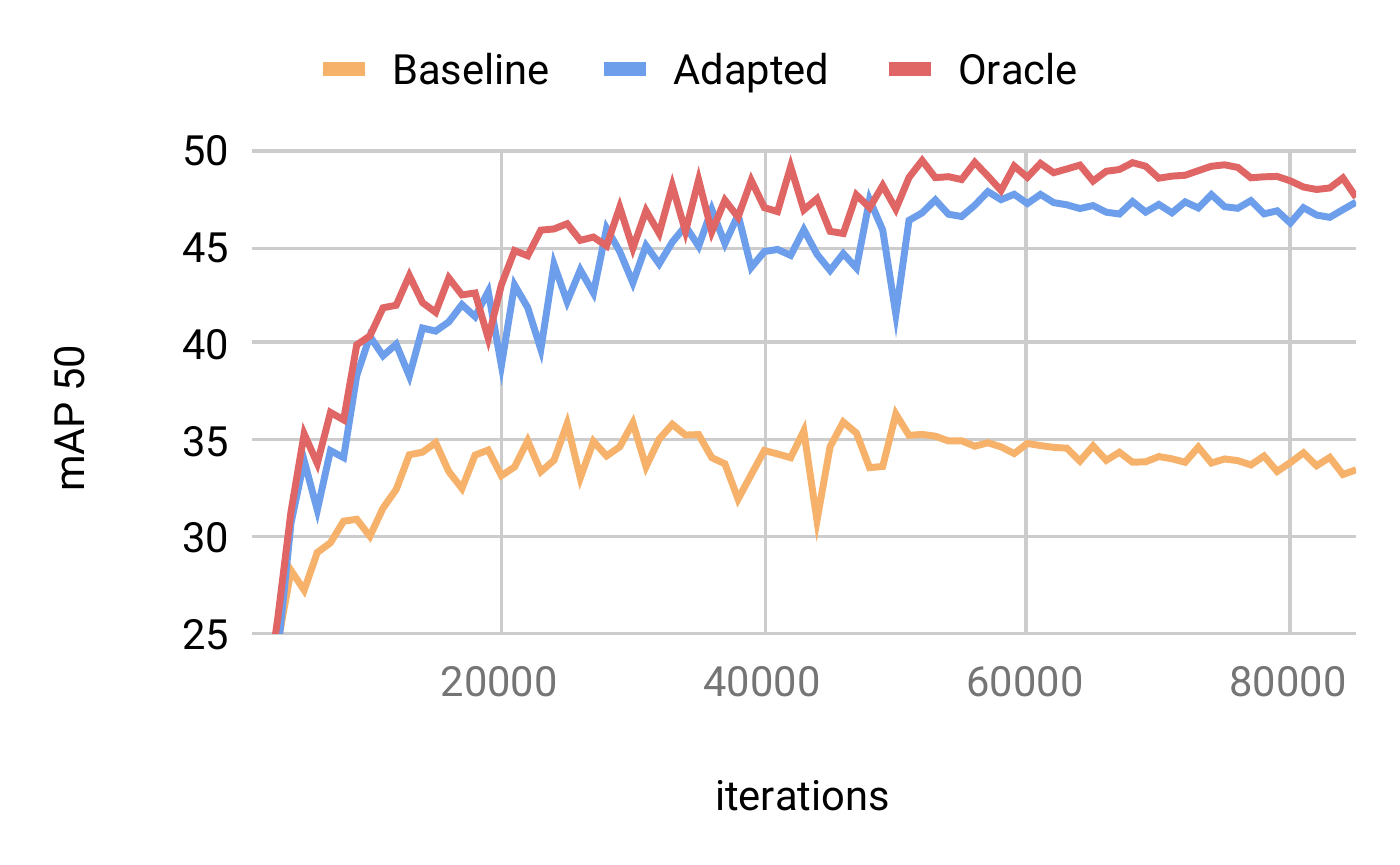}
  \caption{Cityscapes$\rightarrow$Foggy}
  \label{fig:cityscapes2foggycityscapes}
\end{subfigure}%
\begin{subfigure}{.33\textwidth}
  \centering
  \includegraphics[width=\linewidth]{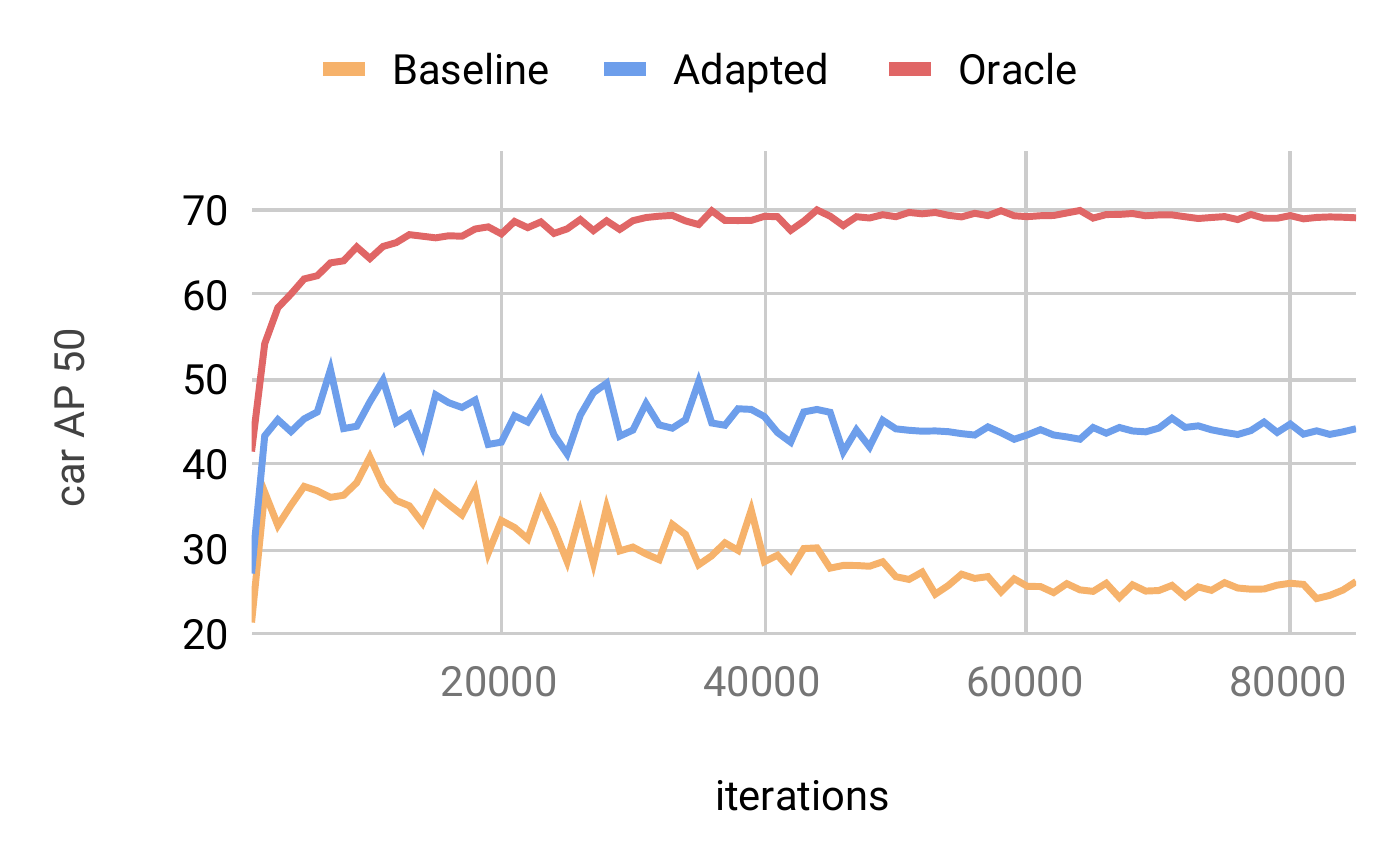}
  \caption{KITTI$\rightarrow$Cityscapes}
  \label{fig:kitti2cityscapes}
\end{subfigure}
\caption{
AP\textsubscript{50} evaluated on the validation split of the target dataset during the training every 1000 iterations for a total 85k iterations. Baseline is a model trained on source only, while the oracle is trained on target only.
}
\label{fig:fig}
\end{figure*}
\begin{table*}[ht]
\centering
\resizebox{.8\textwidth}{!}{%
\begin{tabularx}{1\textwidth}{lY@{}YY@{}YY@{}Y}
\toprule

Method                                      & \multicolumn{2}{c}{\textit{sim2real}} & \multicolumn{2}{c}{\textit{adverse weather}} & \multicolumn{2}{c}{\textit{cross-camera}} \\ \midrule
baseline~\cite{ren2015faster} (ours)          & 34.91             & -                 & 33.48             & -                        & 25.11               & -                   \\ \midrule
DA-Faster-RCNN~\cite{chen2018domain}        & 38.97             & +8.85             & 27.60             & +8.8                     & 38.50               & +8.30               \\
Pseudo-Labeling~\cite{inoue2018cross}       & 39.05             & +7.97             & 33.90             & +2.0                     & 40.23               & +9.13               \\
MTOR~\cite{cai2019exploring}                & 46.60             & +7.20             & 35.10             & +8.2                     & -                   & -                   \\
PDA~\cite{hsu2019progressive}               & -                 & -                 & 36.90             & +10.0                    & 43.90               & +5.70               \\
SCDA~\cite{zhu2019adapting}                 & 43.02             & +9.06             & 33.80             & +7.6                     & 42.50               & +5.10               \\
SWDA~\cite{saito2019strong}                 & 41.50             & +6.90             & 34.30             & +14.0                    & -                   & -                   \\
DM~\cite{kim2019diversify}                  & -                 & -                 & 34.60             & \underline{+16.7}           & -                   & -                   \\
Noisy Labeling~\cite{khodabandeh2019robust} & 42.56             & +11.48            & 36.45             & +4.5                     & 42.98               & +11.88              \\
MAF~\cite{he2019multi}                      & 41.10             & +11.00            & 34.00             & +15.2                    & 41.00               & +10.80              \\ 
CFFA~\cite{zheng2020cross}                  & 43.80             & +8.80            & 38.60             & \underline{\textbf{+17.8}}                    & -               & -              \\ 
\midrule
\textbf{ours}                               & \textbf{50.22}    & \textbf{+15.31}   & \textbf{47.31}    & +13.8                    & \textbf{44.25}      & \textbf{+19.14}     \\
oracle (ours)                                      & 69.03             & +34.12            & 47.56             & +14.1   & 69.03               & +34.12              \\ \bottomrule
\end{tabularx}%
}
\caption{Quantitative analysis of the adaptation efficiency measured in AP\textsubscript{50} and relative improvement w.r.t. the reported baseline. Underlined scores mark higher relative improvements than the \textit{oracle} (a model trained entirely on labeled target data), which is due to diminishing returns.}
\label{tab:extensive-comparison}
\end{table*}
\begin{table}[ht]
\centering
\resizebox{.45\textwidth}{!}{%
\begin{tabular}{@{}cccccl@{}}
\toprule
Experiment                                                                          & Method                        & Baseline & Adapted & Oracle & \multicolumn{1}{c}{Coverage} \\ \midrule
\multirow{2}{*}{\textit{sim2real}}                                                  & MTOR~\cite{cai2019exploring}  & 39.40    & 46.60    & 58.60  & 37.50\%                      \\
                                                                                    & ours                          & 34.91    & 50.22    & 69.03  & \textbf{44.87\%} (+7.37\%)            \\ \midrule
\multirow{2}{*}{\textit{\begin{tabular}[c]{@{}c@{}}adverse\\ weather\end{tabular}}} & CFFA~\cite{zheng2020cross}    & 20.80    & 38.60    & 43.30  & 79.11\%                      \\
                                                                                    & ours                          & 33.48    & 47.31    & 47.56  & \textbf{98.22\%} (+19.11\%)           \\ \midrule
\multirow{2}{*}{\textit{\begin{tabular}[c]{@{}c@{}}cross-\\ camera\end{tabular}}}   & PDA~\cite{hsu2019progressive} & 38.20    & 43.90    & 55.80  & 32.39\%                      \\
                                                                                    & ours                          & 25.11    & 44.25    & 69.03  & \textbf{43.58\%} (+11.19\%)           \\ \bottomrule
\end{tabular}%
}
\caption{Supervised Accuracy Coverage comparison to state-of-the-art methods in various domain adaptation scenarios}
\label{tab:accuracy-coverage}
\end{table}

\noindent\textbf{Transfer from Simulation to Real World (sim2real).}

In this setting we evaluate how well can knowledge from synthetic data be utilized in a real world setting.
The importance of such experimental setting is relevant in scenarios where annotating real data with humans is expensive compared to the cost of 1) generating diverse labeled synthetic images from the simulator and 2) acquiring new input images from the target domain.
For training we use the entire SIM 10K~\cite{johnson2016driving} dataset as the source domain ($\mathcal{X}^s, \mathcal{Y}^s$), while we use the training split of Cityscapes~\cite{cordts2016cityscapes} as the target domain unlabeled images $\mathcal{X}^t$.
We evaluate the final model performance on the validation split of Cityscapes.
Since only the car instances exists in SIM 10k, we report the \textit{car AP\textsubscript{50}}.

Results are shown in the \textit{sim2real} column of Table~\ref{tab:extensive-comparison}, where we compare our performance against previous methods.
As the reported baseline performance (Faster-RCNN~\cite{ren2015faster} trained on source) in several existing works differ significantly, we report the total accuracy and the relative improvement with regards to \textbf{the baseline accuracy reported in each paper} for fair comparison. 
Our method outperforms the previous state-of-the-art approach MTOR~\cite{cai2019exploring} by \textbf{+3.6\%}, and their reported relative improvement (\textbf{7.2\%}) with regards to their baseline is much smaller than ours (\textbf{15.31\%}).
From the perspective of relative improvement our method is also superior to other adaptation techniques: \cite{khodabandeh2019robust} reports a relative improvement of +11.48\% while our method has \textbf{+15.31\%} relative improvement.
The high relative improvement manifests that our accuracy improvements stems from the proposed multilevel knowledge translation approach instead of a stronger baseline model.
In Figure~\ref{fig:sim10k2cityscapes} we show that our technique \textit{covers} more than \textbf{44\%} of the performance gap, meanwhile the baseline begins to over-fit the source domain after 50k iterations.

\noindent\textbf{Adaptation to Adverse Weather (changing weather).}
In this setting, we measure how changing weather scenarios impact the accuracy of a model  trained on a dataset under good weather conditions.
We use labeled images from the training set of Cityscapes~\cite{cordts2016cityscapes} as source ($\mathcal{X}^s, \mathcal{Y}^s$) ,
and adapt our model to unlabeled images from the training split of Foggy Cityscapes~\cite{sakaridis2018semantic} ($\mathcal{X}^t$).
We evaluate AP\textsubscript{50} for all 8 categories in Foggy Cityscapes since it has compatible categories with Cityscapes.
The model trained after 70k iterations is utilized for evaluation.

In the \textit{adverse weather} column of Table~\ref{tab:extensive-comparison}, we show that our method outperforms all prior work in terms of mAP\textsubscript{50}.
In terms of mAP\textsubscript{50}, the previous state-of-the-art -- CFFA~\cite{zheng2020cross} -- has achieved 36.9\% mAP, while our results show a significant \textbf{+8.71\%} improvement.
In terms of relative improvement w.r.t. reported baseline, we again outperform previous works with +13.8\% except MAF~\cite{he2019multi} and CFFA~\cite{zheng2020cross} that used a weaker baseline compared to ours. 
We argue that improvement upon a strong baseline is more convincing. 
To provide further evidence for the difficulty of improving on stronger baselines we have highlighted in the last row of Table~\ref{tab:cityscapes2foggycityscapes} that even the oracle's relative improvement could not exceed MAF~\cite{he2019multi}.
In Figure~\ref{fig:cityscapes2foggycityscapes-bars}, we demonstrate that our baseline performance exceeds DAF~\cite{chen2018domain} and it is on-par with the  accuracy of MAF~\cite{he2019multi} after adaptation.
To illustrate the effectiveness of our approach, we plot the validation accuracy in Figure~\ref{fig:cityscapes2foggycityscapes}, which shows the \textit{coverage} of the performance gap between the baseline and the oracle is above \textbf{98\%}.
A detailed class-specific comparison of AP scores can be found in Sec.~\ref{sup:cityscapes2foggycityscapes}.

\noindent\textbf{Cross-camera adaptation (cross-camera).}
Finally, we show results under a setting where both the source and the target dataset is real, however the input images were captured in different lighting conditions in various environments with different cameras.
We use the entire KITTI~\cite{geiger2013vision} train set as the source domain ($\mathcal{X}^s, \mathcal{Y}^s$), and the unlabeled images of the train set of Cityscapes~\cite{cordts2016cityscapes} as the target domain ($\mathcal{X}^t$).
Following prior studies we report \textit{car AP\textsubscript{50}} scores on the validation set of Cityscapes after 70k iterations.

Our results on cross-camera adaptation are summarized in the \textit{cross-camera} column of Table~\ref{tab:extensive-comparison}.
We again outperform all compared approaches and achieve the highest relative improvement  \textbf{+19.14\%}.
In Figure~\ref{fig:kitti2cityscapes} we show that Multilevel Knowledge Transfer \textit{covers} more than \textbf{43\%} of the performance gap, which manifests the effectiveness of the approach in closing the performance gap caused by domain shifts.

\subsection{Ablation studies and analysis}
\label{subsec:ablation}

In the following, we will discuss the effectiveness of different componenents in our model and how they collaborate to achieve the state-of-the-art performance.
To determine which domain adaptation component plays the most significant role in improving the target domain accuracy in each experimental setting, we have trained different detection models as below.
First we trained the baseline model using the labeled source data only.
Next, we  train models with different combinations of the studied three levels of  domain adaptation (image, feature, output).
The detailed results and comparisons are shown in Table~\ref{tab:ablation}.

\begin{table}[h]
\centering
\resizebox{.5\textwidth}{!}{%
\begin{tabularx}{.7\textwidth}{cccYYY}
\toprule
IMG      & FEA     & OUT     & \textit{sim2real} & \textit{adverse weather} & \textit{cross-camera} \\ \midrule
         &         &                  & 34.91    & 34.16 & 25.11        \\
         &         & \checkmark       & 36.78    & 37.79 & 25.58        \\
         & \checkmark       &         & 43.55    & 43.83 & 42.00        \\
         & \checkmark       & \checkmark       & 48.81        & 43.71 & 42.95        \\
\checkmark        &         &         & 47.73    & 44.33 & 40.23        \\
\checkmark        &         & \checkmark       & 50.00    & 46.09 & 41.81        \\
\checkmark        & \checkmark       &         & 43.14    & \textbf{47.37} & 43.74        \\
\checkmark        & \checkmark       & \checkmark       & \textbf{50.22}        & \textbf{47.31} & \textbf{44.25}        \\ \midrule
\multicolumn{3}{c}{oracle}   & 69.03    & 47.56 & 69.03        \\ \bottomrule
\end{tabularx}%
}
\caption{Target accuracy (mAP\textsubscript{50}) reported after 70k training iterations under experimental settings described in~\ref{subsec:exp-results}.}
\label{tab:ablation}
\end{table}

\begin{table}[h]
\centering
\resizebox{.45\textwidth}{!}{%
\begin{tabularx}{.73\textwidth}{cYYY}
\toprule
FID                             & \textit{sim2real} & \textit{adverse weather} & \textit{cross-camera} \\ \midrule
source $\leftrightarrow$ target & 102.86   & 40.49 & 81.01        \\
translated $\leftrightarrow$ target    & 61.88    & 36.94 & 75.06        \\ \bottomrule
\end{tabularx}%
}
\caption{FID~\cite{heusel2017gans} between training image distributions.}
\label{tab:fid}
\end{table}

It is important to notice that the contribution of each level of adaptation differs in different experimental settings and exhibits inconsistent behaviours. 
We attribute these inconsistencies to the different domain shifts imposed by different settings.
In Table~\ref{tab:fid}, we report the Fréchet Inception Distances (FID)~\cite{heusel2017gans} between the input distributions in each experimental setting.
The first row lists the distances between the distribution of original source and target images.
In the second row, we list the distances between the distribution of translated source images and the target images. 

\noindent\textbf{Single component analysis.}
Image level adpatation boosts the performance more than feature level adaptation when the domain discrepancy between source and target is large -- \textit{sim2real} ({+12.82\%} vs.\ {+8.64\%} )
and they yield similar performance improvement when the domain dicrepancy is small -- \textit{changing weather} ({+10.17\%} vs.\ {+9.67\%}) and  \textit{cross-camera} ({+15.12\%} vs.\ {+16.89\%}). In constrast, the output level domain adaptation alone doesn't benefit the performance a lot influenced by the quality of the teacher network.

\noindent\textbf{Component combination analysis.}
We observe combining two components does not always bring performance improvement in comparison to the single component baseline. In the \textit{sim2real} scenario, the performance drops to 43.14\% (similar to feature-level only) when image-level and feature-level adaptation are adopted together. This is potentially caused by imperfect translation results from image to image translation which is the most difficult in \textit{sim2real} scenario. 
Output-level adaptation consistently improves the  performance when combined with other level of domain adaptation approaches in all  settings.  
Overall, our proposed framework incorporating all three levels of adaptation achieves the best performance.

\subsection{Dataset statistics}
\label{sup:datasets}
\subsubsection{Cityscapes}
Cityscapes~\cite{cordts2016cityscapes} is a dataset for autonomous driving experiments in urban settings where images were captured with an on-board device.
It contains a train and validation split with 2,975 and 500 images, respectively.
The training set contains 52088 instances, namely 
\textit{person} (17,910),
\textit{rider} (1,778),
\textit{car} (26,957),
\textit{truck} (484),
\textit{bus} (380),
\textit{train} (168),
\textit{motorcycle} (737),
\textit{bicycle} (3,674).
Since the dataset main purpose is to provide a benchmark for semantic and instance segmentation we cannot directly use its labels nor the corresponding evaluation tool-kit for measuring object-detection, therefore we generate tight bounding boxes for instances following prior approaches and use the COCO~\cite{lin2014microsoft} AP evaluation tool implemented in Detectron2~\cite{wu2019detectron2}.
Under settings where Cityscapes is used as \textit{source} we use the training split.
Under settings where Cityscapes is \textit{target} we use the training split, but without labels.
For evaluation we always use the validation set.

\subsubsection{SIM 10K.}
SIM 10K~\cite{johnson2016driving} has 10,000 synthetic images captured from the video game \textit{Grand Theft Auto V}.
Bounding boxes of 58,701 instances are provided in PASCAL VOC~\cite{pascal-voc-2012} format of \textit{car} (57,776), \textit{person} (4), \textit{motorbike} (921) categories.
Previous papers report using 58,701 car instances, however only 57,776 instances of the provided instances belong to the \textit{car} category.
Also, according to the standard Faster-RCNN~\cite{ren2015faster} settings, we do not use images without instances, which effectively reduces the size of the training set to 9975 images.
During training we use the whole dataset and do not use an arbitrary validation split for hyperparameter tuning.

\subsubsection{Foggy Cityscapes.}
Foggy Cityscapes~\cite{sakaridis2018semantic} is an artificial dataset created to measure performance of models in adverse weather conditions by adding synthetic fog to the real images in Cityscapes~\cite{cordts2016cityscapes}.
This makes compatible to compare category AP scores of models trained on Cityscapes.
The dataset contains three different scenarios with decreasing level of visibility of objects.
We use all three levels of visibility both for training and validation, effectively multiplying the available number of images and annotations by 3.
Under settings where Foggy Cityscapes is \textit{target} we use the training split without labels and for evaluation we use the validation set.

\subsubsection{KITTI.}
KITTI~\cite{geiger2013vision} provides a dataset for autonomous driving in diverse real-world traffic scenarios ranging from rural areas to inner-city environments.
The dataset contains 7481 images with 51,865 instances in total of 9 categories, namely
\textit{car} (28,742),
\textit{van} (2,914),
\textit{truck} (1,094),
\textit{pedestrian} (4,487),
\textit{person sitting} (222),
\textit{cyclist} (1,627),
\textit{tram} (511),
\textit{miscallenous} (973),
\textit{don't care} (11,295).
For the sake of simplicity we discard \textit{don't care} regions from both training evaluation.
Both under settings where KITTI is used as \textit{source} and \textit{target} we use the whole dataset and do not use an arbitrary split for acquiring a validation set to tune hyperparameters.

\subsection{Hard vs. Soft Labels}
\label{sup:soft-labels}
\begin{table}[ht]
\centering
\resizebox{.45\textwidth}{!}{%
\begin{tabular}{@{}rcc@{}}
\toprule
Temperature & student source mAP & student target mAP \\ \midrule
0.5         & 42.66                   & 40.14                   \\
1.0         & 45.01                   & 43.19                   \\
1.1         & 45.21                   & 42.82                   \\
2.0         & 48.49                   & 45.02                   \\
5.0         & 48.93                   & 45.99                   \\
10.0        & 49.00                   & 45.99                   \\
20.0        & 50.51                   & 47.20                   \\ \hline
\textbf{hard-label}  & \textbf{51.23}                   & \textbf{47.30}
\\ \bottomrule
\end{tabular}%
}
\caption{
Quantitative comparison of treating pseudo-labels $\mathcal{Y}^p$ as ground truth.
For the hard-label experiment $\alpha=1$, otherwise $\alpha=\frac{1}{2}$.
}
\label{tab:soft-labels}
\end{table}
We compare the effect of treating teacher predictions as hard-labels versus soft-labels (as introduced in~\cite{hinton2015distilling}) and conclude that treating $y^p$ as hard-labels is favorable in terms of target accuracy.
In our implementation only the classification head's cross-entropy objective function is modified as follows:
\begin{equation*}
    \sum_i \alpha p_i \log q_i + (1-\alpha) \hat{p}_i \log q_i
\end{equation*}
where $p_i$ is the hard-label and $\hat{p}_i$ is the soft-label with temperature parameter $T$, defined by:
\begin{equation*}
    \hat{p}_i = \frac{exp(z_i/T)}{\sum_j exp(z_j/T)}
\end{equation*}

We observe that in settings where $\alpha=0$, the training of the student model becomes unstable resulting in close to 0.0 mAP scores.
Experimental results using \textit{sim2real} adaptation settings are shown in Table~\ref{tab:soft-labels}.
We show that by increasing the temperature $T$ along fixed $\alpha$ parameter the effect of soft-labels becomes less dominant during in the training and the accuracy converges to the hard-label $\alpha=1$ setting.

\subsection{Importance of multi-modal image translation}
\label{sup:cyclegan}
\begin{table}[h]
\centering
\resizebox{.45\textwidth}{!}{%
\begin{tabular}{@{}cccc@{}}
\toprule
image-to-image model & \textit{sim2real} & \textit{adverse weather} & \textit{cross-camera} \\ \midrule
CycleGAN             & 46.82             & \textbf{44.71}           & 34.67                 \\
MUNIT                & \textbf{47.43}    & 44.33                    & \textbf{40.23}        \\ \bottomrule
\end{tabular}%
}
\caption{Qualitative comparison of target accuracy achieved by different image translation models.}
\label{tab:munit-vs-cyclegan}
\end{table}
In Table~\ref{tab:munit-vs-cyclegan}, we provide experimental results on the importance of covering the multi-modality present in the target dataset while performing image to image translation.
CycleGAN~\cite{zhu2017unpaired} allows only a deterministic translation, therefore it provides a \textit{one-to-one} mapping from source to target images.
On the other hand, MUNIT~\cite{huang2018multimodal} allows us to sample different style vectors from a normal distribution to control the appearance of the result of the image translation, providing a \textit{one-to-many} mapping between source and target images.
We observe that in 2 out of 3 domain adaptation scenarios using multi-modal image translation improves the target accuracy. 
In the \textit{adverse weather} scenario is an example of diminishing returns, since converting \textit{Cityscpes}~\cite{cordts2016cityscapes} to the synthetic \textit{Foggy Cityscapes}~\cite{sakaridis2018semantic} target domain is less challenging than the other two settings  (see Table~\ref{tab:fid}), therefore CycleGAN~\cite{zhu2017unpaired} performance is on par with MUNIT~\cite{huang2018multimodal}.

\subsection{Pseudo-Label performance sensitivity to threshold hyper-parameter}
\label{sup:teacher-threshold}
\begin{table}[t]
\centering
\resizebox{.45\textwidth}{!}{%
\begin{tabular}{@{}cccc@{}}
\toprule
teacher threshold & number of pseudo-labels & student source mAP & student target mAP \\ \midrule
0.0               & 296k                    & 50.10              & 34.27              \\
0.1               & 219k                    & 50.20              & 36.81              \\
0.2               & 160k                    & 49.92              & 36.8               \\
0.3               & 132k                    & 50.28              & 37.64              \\
0.4               & 115k                    & 50.40              & \textbf{38.01}     \\
0.5               & 103k                    & \textbf{50.43}     & 37.98              \\
0.6               & 93k                     & 49.06              & 37.14              \\
0.7               & 85k                     & 49.62              & 36.34              \\
0.8               & 77k                     & \textbf{50.43}     & 36.93              \\
0.9               & 68k                     & 49.58              & 35.18              \\ \bottomrule
\end{tabular}%
}
\caption{Qualitative comparison of different cut-off parameters for selecting pseudo-labels.}
\label{tab:teacher-threshold}
\end{table}
Applying the teacher network to target images yields a set of predictions which we reuse as pseudo-labels for training the student network.
However, it is not trivial whether we should treat every prediction as a correct label or not.
In our experiments we used a predetermined threshold applied on the per-instance  \textit{objectness-scores} (see Faster-RCNN~\cite{ren2015faster} for details) to select which predictions to keep.
In Table~\ref{tab:teacher-threshold}, we show experimental results on the sensitivity of the pseudo-labeling step with regards to the target accuracy measured on the \textit{adverse weather} adaptation scenario.

\subsection{Class specific comparison for the Adverse Weather experiment}
\label{sup:cityscapes2foggycityscapes}
Since the target dataset, Foggy Cityscapes~\cite{sakaridis2018semantic} is artificially generated from the source dataset, Cityscapes~\cite{cordts2016cityscapes} the categories are retained.
The class specific comparison can be found in Table~\ref{tab:cityscapes2foggycityscapes}.

\begin{table}[h]
\centering
\resizebox{.48\textwidth}{!}{%
\begin{tabularx}{\textwidth}{@{}lc *{11}{Y}@{}}
\toprule
Method                                    & person\textsubscript{50} & rider\textsubscript{50} & car\textsubscript{50} & truck\textsubscript{50} & bus\textsubscript{50} & train\textsubscript{50} & mcycle\textsubscript{50} & bicycle\textsubscript{50} & \multicolumn{2}{c}{mAP\textsubscript{50}} \\ \hline
Faster-RCNN~\cite{ren2015faster}          & 36.9                     & 44.0                    & 47.9                  & 18.3                    & 34.5                  & 21.2                    & 25.3                     & 39.8                      & 33.48               & -                   \\ \hline
DA-Faster-RCNN~\cite{chen2018domain}      & 25.0                     & 31.0                    & 40.5                  & 22.1                    & 35.3                  & 20.2                    & 20.0                     & 27.1                      & 27.60               & +8.8                \\
Pseudo-Labeling~\cite{inoue2018cross}     & 31.9                     & 39.9                    & 48.0                  & 25.1                    & 39.9                  & 27.2                    & 25.0                     & 34.1                      & 33.90               & +2.0                \\
MTOR~\cite{cai2019exploring}              & 30.6                     & 41.4                    & 44.0                  & 21.9                    & 38.6                  & 40.6                    & 28.3                     & 35.6                      & 35.10               & +8.2                \\
PDA~\cite{hsu2019progressive}             & 36.0                     & 45.5                    & 54.4                  & 24.3                    & 44.1                  & 25.8                    & 29.1                     & 35.9                      & 36.90               & +10.0               \\
SCDA~\cite{zhu2019adapting}               & 33.5                     & 38.0                    & 48.5                  & 26.5                    & 39.0                  & 23.3                    & 28.0                     & 33.6                      & 33.80               & +7.6                \\
SWDA~\cite{saito2019strong} & 36.2 & 35.3 & 43.5 & 30.0 & 29.9 & 42.3 & 32.6 & 24.5 & 34.3 & +14.0 \\
DM~\cite{kim2019diversify}                & 30.8                     & 40.5                    & 44.3                  & 27.2                    & 38.4                  & 34.5                    & 28.4                                                 & 32.2                     & 34.60                   & +16.7 \\
Noisy Labeling~\cite{khodabandeh2019robust} & 35.1                     & 42.2                    & 49.2                  & 30.1                    & 45.3                  & 27.0                    & 26.9                     & 36.0                      & 36.45               & +4.5                \\
MAF~\cite{he2019multi}                    & 28.2                     & 39.5                    & 43.9                  & 23.8                    & 39.9                  & 33.3                    & 29.2                     & 33.9                      & 34.00               & +15.2 \\
CFFA~\cite{zheng2020cross} & 34.0 & 40.9 & 52.1 & 30.8 & 43.2 & 29.9 & 34.7 & 37.4 & 38.6 & \textbf{+17.8} \\ \hline
ours                                      & \textbf{43.5}            & \textbf{52.0}           & \textbf{63.2}         & \textbf{34.7}           & \textbf{52.7}         & \textbf{45.8}           & \textbf{37.1}            & \textbf{49.4}             & \textbf{47.31}      & +13.8               \\ \hline
oracle                                    & 44.4                     & 51.5                    & 66.0                  & 38.0                    & 55.0                  & 37.8                    & 38.6                     & 49.2                      & 47.56               & \textcolor{red}{+14.1}               \\ \bottomrule
\end{tabularx}%
}
\caption{
Cityscapes $\rightarrow{}$ Foggy Cityscapes.
The unsupervised domain adaptation performance is on par with the network trained on the target (covers 98\% of the performance gap).
As we highlighted CFFA~\cite{zheng2020cross} achieves the highest respective improvement, however their baseline provides larger room for improvement (see Figure~\ref{fig:cityscapes2foggycityscapes-bars}). 
}
\label{tab:cityscapes2foggycityscapes}
\end{table}

\subsection{Fair Comparison}
\label{sup:fair-comparison}

\noindent\textbf{Detector backbone and optimization.}
Prior works implement the object detector, Faster-RCNN~\cite{ren2015faster} using different backbones. VGG16 is used in~\cite{chen2018domain,zhu2019adapting,he2019multi,hsu2019progressive,kim2019diversify}, Inception V2 in~\cite{khodabandeh2019robust}, and~\cite{cai2019exploring} uses 50/152-layer ResNet~\cite{he2016deep} for different settings. 

\noindent\textbf{Evaluation.}
The choice of evaluation tools is missing from existing works.
This is problematic because the mean Average Precision (mAP) metric is sensitive to the underlying implementation.
Using the pre-trained networks from Detectron2~\cite{wu2019detectron2} for VOC 2007 and the default VOC evaluation tool the reported mAP and mAP50 performance is 51.9 and 80.3, while evaluating the same network on the same dataset with the COCO evaluation tool the corresponding scores are 46.9 and 76.2, resulting in a difference of 5.0\% and 4.1\% between the two measurements. This ``relative improvement'' from choosing an appropriate evaluation tool would be on-par with the improvement of~\cite{inoue2018cross,khodabandeh2019robust,zhu2019adapting} (+1.98\%, +4.53\%, +5.1\%).

\section{Conclusion}
In this work we have addressed the problem of cross-domain object detection.
After conducting extensive studies on previous approaches, we categorized different components into 3 classes: image, feature and output level domain adaptation.
Our proposed method is the first to successfully employ all levels of domain adaptation.
Our method consists of 3 stages: 
1) translate labeled source images to make them appear similar to target images 
2) train a teacher network on translated images while aligning its features to the target domain 
3) use the teacher model to generate pseudo-labels, then train a student model on the translated source images and the pseudo-labeled target images.
Compared to recent state-of-the art works, our method uses simple adaptation techniques at each stage and still outperforms complex algorithms on every commonly used benchmark.

\bibliographystyle{unsrt}
\bibliography{related}

\end{document}